\documentclass[letterpaper]{article} 
\usepackage{aaai2026}  
\usepackage{times}  
\usepackage{helvet}  
\usepackage{courier}  
\usepackage[hyphens]{url}  
\usepackage{graphicx} 
\usepackage{natbib}  
\usepackage{caption} 
\usepackage{algorithm}
\usepackage{algorithmic}

\usepackage{newfloat}
\usepackage{listings}
\DeclareCaptionStyle{ruled}{labelfont=normalfont,labelsep=colon,strut=off} 
\floatstyle{ruled}
\newfloat{listing}{tb}{lst}{}
\floatname{listing}{Listing}
\title{Evaluating Large Language Models on Rare Disease Diagnosis: A Case Study using House M.D.}
\author{
    Arsh Gupta \textsuperscript{\rm 1}, 
    Ajay Narayanan Sridhar \textsuperscript{\rm 1}, 
    Bonam Mingole \textsuperscript{\rm 1},     
    Amulya Yadav \textsuperscript{\rm 1} \\
}
\affiliations{
    \textsuperscript{\rm 1}The Pennsylvania State University \\

    abg6210@psu.edu, afs6372@psu.edu, bjm6940@psu.edu, amulya@psu.edu
}

\begin{document}

\maketitle

\begin{abstract}
Large language models (LLMs) have demonstrated capabilities across diverse domains, yet their performance on rare disease diagnosis from narrative medical cases remains underexplored. We introduce a novel dataset of 176 symptom-diagnosis pairs extracted from \textit{House M.D.}, a medical television series validated for teaching rare disease recognition in medical education. We evaluate four state-of-the-art LLMs such as GPT 4o mini, GPT 5 mini, Gemini 2.5 Flash, and Gemini 2.5 Pro on narrative-based diagnostic reasoning tasks. Results show significant variation in performance, ranging from 16.48\% to 38.64\% accuracy, with newer model generations demonstrating a 2.3$\times$ improvement. While all models face substantial challenges with rare disease diagnosis, the observed improvement across architectures suggests promising directions for future development. Our educationally validated benchmark establishes baseline performance metrics for narrative medical reasoning and provides a publicly accessible evaluation framework for advancing AI-assisted diagnosis research.
\end{abstract}

\section{Introduction}

The rapid advancement of large language models (LLMs) has opened new opportunities for applying natural language understanding to complex domains requiring reasoning under uncertainty \cite{jerrentrup2018house, mechler2020house, sanges2020roleplay}. In healthcare, accurate diagnosis from patient-reported symptoms remains a critical challenge where LLMs could provide value for clinical decision support and medical education. However, evaluating LLM performance in medical diagnosis faces significant barriers due to privacy constraints, limited dataset availability, and the formalized nature of existing medical datasets.

Medical television dramas, particularly \textit{House M.D.}, offer a unique solution by providing rich, case-based scenarios where symptoms and diagnostic outcomes are clearly described in narrative form~\cite{jerrentrup2018house,mechler2020house,cambra2021medical}. The educational value of \textit{House M.D.} has been well-established, with successful use in medical education to teach about rare diseases and clinical reasoning~\cite{jerrentrup2018house,mechler2020house}. Research shows that 49.6\% of health sciences students watch medical dramas regularly, and these shows effectively convey medical knowledge~\cite{cambra2021medical}.

In this work, we evaluate LLM performance on symptom-to-diagnosis prediction using a novel dataset of 176 cases constructed from \textit{House M.D.} episodes. Each case pairs symptom descriptions with corresponding disease diagnoses, demanding that models identify medical clues within narrative descriptions, closely mirroring clinical practice. Our dataset is publicly available on Kaggle\footnotemark

\footnotetext{
\url{https://bit.ly/4p9ltW8}
}

Our work addresses three challenges: (i) general-purpose LLMs lack specialization for narrative medical reasoning, (ii) available datasets rarely capture narrative-driven diagnostic structure, and (iii) limited dataset sizes raise generalizability concerns. We evaluate multiple state-of-the-art LLMs like GPT 5 Mini~\cite{openai2025gpt5mini}, GPT 4o Mini~\cite{openai2024gpt4o-mini}, Gemini 2.5 Flash~\cite{google2025gemini2_5flash}, and Gemini 2.5 Pro~\cite{google2025gemini2_5pro} to assess diagnostic reasoning capabilities across different architectures. Our experiments reveal modest performance on rare disease diagnostic tasks, highlighting domain-specific challenges in medical reasoning from narrative cases.

Our contributions are: (1) a novel dataset of symptom-disease mappings from publicly available narrative cases reflecting clinical reasoning patterns, and (2) comprehensive baseline evaluation across multiple LLM architectures demonstrating current limitations in narrative medical reasoning.

Results establish clear performance baselines across different model families, indicating that narrative-based rare disease diagnosis remains a significant challenge for current LLMs. These findings suggest that educationally validated medical narratives can serve as valuable benchmarks for measuring LLM diagnostic capabilities, establishing a foundation for future research in AI-assisted medical diagnosis and highlighting the need for domain-specific adaptations.

\section{Related Work}

Medical television dramas have gained recognition as effective educational tools with structured narrative formats suitable for both human learning and machine learning applications. Jerrentrup et al. \cite{jerrentrup2018house} demonstrated that \textit{House M.D.} can be successfully integrated into medical curricula for teaching rare diseases and complex diagnostic scenarios. Cambra-Badii et al. \cite{cambra2021medical} found that 49.6\% of health sciences students regularly watch medical dramas, with \textit{House M.D.} among the most popular, and that these shows effectively teach bioethical and professional practice issues.

Sanges et al.~\cite{sanges2020roleplay} and Sarrafpour et al.~\cite{sarrafpour2021simulation} validated case-based learning for rare disease education, showing significant improvements in students' diagnostic recognition abilities. These findings suggest that structured symptom-diagnosis relationships in narrative medical content may provide valuable evaluation data for AI systems reasoning through clinical presentations.

Beyond educational validation, studies have highlighted the unique coverage of rare diseases in medical television. Mechler et al.~\cite{mechler2020house} analyzed orphan diseases featured in \textit{House M.D.}, showing the series frequently presents rare diseases seldom encountered in training, making it valuable for both medical students and AI diagnostic evaluation.

Schaefer and von Hirschhausen~\cite{schaefer2021media} demonstrated that medical television helps viewers recognize symptoms and seek appropriate consultation. Furman and Clayton~\cite{furman2021genetics} analyzed genetic concepts in medical shows, highlighting that narrative case presentations closely parallel diagnostic reasoning challenges, suggesting clear utility for evaluating automated diagnostic systems.

Recent advances emphasize simulation and case-based learning approaches mirroring structured narratives in medical television. Rattani et al.~\cite{rattani2021narrative} identified narrative methods as particularly effective for complex scenarios, while Rasul et al.~\cite{rasul2024realworld} validated real-scenario approaches in medical education. These studies demonstrate that well-structured narrative cases effectively bridge theoretical knowledge and practical application, indicating their potential for evaluating LLMs on diagnostic reasoning patterns.

While existing literature supports the educational value of medical television for human learners, limited research explores its application to machine learning systems. Our work addresses this gap by systematically extracting symptom-diagnosis pairs from \textit{House M.D.} episodes and evaluating LLM diagnostic capabilities across multiple architectures, extending established educational value of medical narratives to AI evaluation.

\section{Dataset Creation}

We constructed our dataset from the publicly available \textit{House M.D.} wiki (\url{https://house.fandom.com}), extracting narrative content from all 176 episodes across eight seasons. The data extraction involved three stages: (1) web scraping using \texttt{BeautifulSoup}~\cite{richardson2024beautifulsoup4}, which allows for robust parsing of HTML content, (2) structured prompt generation where each evaluated model (GPT-4o~mini~\cite{openai2024gpt4o-mini}, GPT-5~Mini~\cite{openai2025gpt5mini}, Gemini~2.5~Flash~\cite{google2025gemini2_5flash}, Gemini~2.5~Pro~\cite{google2025gemini2_5pro} independently transforms raw narrative episodes into standardized medical case formats, and (3) quality filtering to ensure clinical detail, diagnostic clarity, and alignment with real-world medical reasoning~\cite{jerrentrup2018house,cambra2021medical,sanges2020roleplay}.

Our final dataset consists of 176 symptom-diagnosis pairs spanning diverse medical specialties, with emphasis on complex diagnostic scenarios requiring multi-step reasoning. The complete dataset and model evaluation results are publicly available on Kaggle and GitHub.\footnotemark

\footnotetext{
Code: \url{https://bit.ly/481OSuD}, 
\url{https://bit.ly/4oCxqnj};\\
Dataset: \url{https://bit.ly/4p9ltW8};\\
Results: \url{https://bit.ly/3JYGEeG}, 
\url{https://bit.ly/47FHXs1}
}

\textbf{Educational Validation:} Jerrentrup et al.~\cite{jerrentrup2018house} demonstrated successful integration of \textit{House M.D.} into medical curricula for teaching rare disease recognition.

\textbf{Rare Disease Coverage:} Mechler et al.~\cite{mechler2020house} found the show frequently features orphan diseases underrepresented in traditional medical datasets, addressing a gap in medical AI evaluation benchmarks.

\textbf{Clinical Realism:} Despite dramatic elements, the show employs medical consultants to ensure clinical accuracy and follows a consistent diagnostic framework mirroring practice.

\textbf{Accessibility:} Unlike proprietary medical datasets, \textit{House M.D.} content is publicly available, enabling reproducible research while avoiding ethical constraints of real patient data.

\textbf{Narrative Context:} The narrative format preserves temporal symptom progression, demographics, and clinical decision-making that structured datasets often lose~\cite{cambra2021medical,schaefer2021media}.

While our dataset reflects limitations of fictional content, including dramatic exaggeration and complex case focus, these characteristics may benefit evaluation by providing challenging edge cases that test model robustness. The educational validation of \textit{House M.D.} by medical professionals provides confidence that extracted scenarios contain clinically meaningful information suitable for AI evaluation~\cite{cambra2021medical,schaefer2021media}.

\section{Methodology}

We designed a straightforward evaluation pipeline to assess LLM performance on symptom-to-diagnosis prediction tasks across multiple model architectures, consisting of prompt construction, model inference, and evaluation metrics.

\begin{figure}[h]
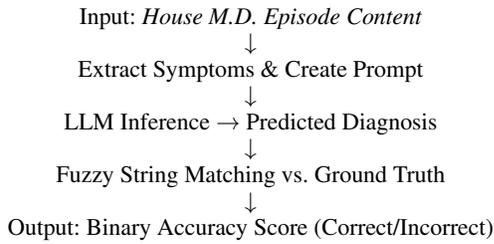

\centering
\small
\begin{tabular}{c}
Input: \textit{House M.D. Episode Content} \\
↓ \\
Extract Symptoms \& Create Prompt \\
↓ \\
LLM Inference → Predicted Diagnosis \\
↓ \\
Fuzzy String Matching vs. Ground Truth \\
↓ \\
Output: Binary Accuracy Score (Correct/Incorrect) \\
\end{tabular}
\caption{Evaluation pipeline for LLM-based diagnosis}
\label{fig
:pipeline}
\end{figure}

\begin{table}[h]
\centering
\small
\begin{tabular}{|p{1.3cm}|p{2.5cm}|p{3cm}|}
\hline
\textbf{Stage} & \textbf{Process} & \textbf{Example} \\ \hline
Exact Match & Prediction in ground truth & \textit{Lupus} in \textit{Systemic Lupus} \\ \hline
Fuzzy Match & Token similarity (thresh. = 0.8) & \textit{Sarcoidosis} $\approx$ \textit{Sarcoid} (0.89) \\ \hline
Classif. & Binary outcome & Correct / Incorrect \\ \hline
\end{tabular}
\caption{Fuzzy matching workflow}
\label{tab:fuzzy}
\end{table}
\subsection{Model Selection and Configuration}
We evaluated four state-of-the-art LLMs: GPT-4o Mini, GPT-5 Mini, Gemini 2.5 Flash, and Gemini 2.5 Pro. This selection spans different model families (OpenAI and Google) and capability levels, enabling assessment of diagnostic reasoning across various architectures and training approaches.

The model was configured with standard parameters: temperature set to 0.0 to ensure deterministic outputs, maximum token length of 1500 to accommodate detailed diagnostic reasoning, and no additional system prompts beyond the diagnostic instruction to avoid introducing bias toward specific medical frameworks.

\subsection{Prompt Design and Inference}
Our prompts follow a structured medical case presentation format designed to simulate realistic clinical scenarios. Each prompt contains patient demographic information, symptom descriptions with temporal progression, relevant medical history, and initial diagnostic workup results. The prompts explicitly request a single primary diagnosis while encouraging the model to provide supporting reasoning.

For each case, models generate diagnostic responses in a single-pass approach without iterative refinement. Model responses were collected systematically across all 176 cases with consistent experimental conditions.

\begin{table}[h]
\centering
\scriptsize
\begin{tabular}{|p{2.5cm}|p{2.5cm}|p{1.8cm}|p{2.5cm}|c|}
\hline
\textbf{Prompt} & \textbf{Ground Truth Disease} \\ \hline
\textit{"A 27-year-old male presents to the ER after an episode of acute aphonia and syncope during his wedding ceremony. Initially, malingering was suspected, but he subsequently developed a productive cough, cyanosis, and was found to have a pleural effusion. His workup revealed a positive mononucleosis test, which is atypical for his age......"} & 
Arnold-Chiari malformation \\ \hline
\end{tabular}
\caption{Example evaluation case from Gemini 2.5 Pro showing only a portion of the Prompt and Ground Truth Disease.}
\label{tab:example_case}
\end{table}

\subsection{Evaluation Metrics}
We evaluated predictions using fuzzy string matching against ground truth diagnoses, addressing the challenge that medical conditions have multiple valid names. Our algorithm employs Python's \texttt{SequenceMatcher} with a 0.8 similarity threshold, performing exact substring matching first, then token-wise fuzzy comparison. Final accuracy is computed as the proportion of correctly classified cases, providing clear performance benchmarks while accommodating medical terminology ambiguity.

\subsection{Limitations}

Our methodology has acknowledged limitations. Fuzzy matching may miss semantically equivalent diagnoses using substantially different terminology, and the binary metric may not capture partial credit for related diagnoses. However, our approach provides a systematic and reproducible framework for evaluating LLM diagnostic performance across multiple architectures.

\section{Results}

\subsection{Overall Performance}
We evaluated four state-of-the-art LLMs on our \textit{House M.D.} diagnostic dataset using fuzzy string matching. Table~\ref{tab:model_performance} summarizes the accuracy across all models.

\begin{table}[h]
\centering
\small
\begin{tabular}{|l|c|c|}
\hline
\textbf{Model} & \textbf{Correct} & \textbf{Accuracy (\%)} \\ \hline
GPT-4o-mini & 29/176 & 16.48 \\ \hline
Gemini 2.5 Flash & 58/176 & 32.95 \\ \hline
GPT-5-mini & 65/176 & 36.93 \\ \hline
Gemini 2.5 Pro & 68/176 & 38.64 \\ \hline
\end{tabular}
\caption{Diagnostic accuracy across LLM models}
\label{tab:model_performance}
\end{table}

Performance varied significantly across model architectures, with Gemini 2.5 Pro achieving the highest accuracy at 38.64\%, followed by GPT-5 Mini at 36.93\%, Gemini 2.5 Flash at 32.95\%, and GPT-4o Mini at 16.48\%. Despite these differences, all models demonstrated substantial challenges with rare disease diagnostic reasoning.

\subsection{Performance Analysis}

Performance varied not only across models but also across seasons, as shown in Table~\ref{tab:season_performance}. Season 1 achieved the highest accuracy at 56.52\%, while Season 5 showed the lowest at 20.83\%. This variation suggests that diagnostic complexity varies throughout the series, with later seasons potentially featuring more challenging rare disease cases. However, the relatively strong performance in Season 8 (52.38\%) indicates that temporal progression alone does not fully explain accuracy differences rather case-specific diagnostic complexity appears to be the primary driver.

\begin{table}[h]
\centering
\small
\begin{tabular}{|l|c|c|c|}
\hline
\textbf{Season} & \textbf{Episodes} & \textbf{Correct} & \textbf{Accuracy (\%)} \\ \hline
Season 1 & 23 & 13 & 56.52 \\ \hline
Season 2 & 24 & 7 & 29.17 \\ \hline
Season 3 & 24 & 8 & 33.33 \\ \hline
Season 4 & 16 & 7 & 43.75 \\ \hline
Season 5 & 24 & 5 & 20.83 \\ \hline
Season 6 & 21 & 8 & 38.10 \\ \hline
Season 7 & 23 & 9 & 39.13 \\ \hline
Season 8 & 21 & 11 & 52.38 \\ \hline
\end{tabular}
\caption{Per-season diagnostic accuracy for Gemini 2.5 Pro}
\label{tab:season_performance}
\end{table}

Across all models, performance was better on common conditions with distinctive symptom presentations (meningitis, myocardial infarction, pulmonary embolism). All models struggled with rare diseases (neurocysticercosis, Erdheim-Chester disease), multi-system autoimmune disorders (systemic lupus erythematosus, sarcoidosis), and toxicological cases requiring integration of exposure history with clinical presentation.

The performance gap between models suggests that architectural differences and training approaches significantly impact diagnostic reasoning capabilities. GPT-5-mini and Gemini 2.5 Pro's superior performance indicates that newer model generations with enhanced reasoning capabilities show meaningful improvements over earlier versions, though substantial limitations remain.

\subsection{Implications}

These results establish important baseline performance metrics for narrative-based rare disease diagnosis and demonstrate that current LLMs show promising capabilities in medical reasoning tasks. The improvement from GPT-4o Mini (16.48\%) to Gemini 2.5 Pro (38.64\%) indicates that the field is making meaningful progress toward clinically useful AI diagnostic systems. While absolute accuracy levels indicate room for improvement, it is important to contextualize these results: our benchmark exclusively features diagnostically challenging cases that often puzzle expert physicians, representing a substantially harder evaluation task than typical medical AI benchmarks. The ability to correctly diagnose nearly 40\% of these exceptionally difficult cases demonstrates meaningful medical reasoning capabilities and establishes a solid foundation for future improvements through domain-specific fine-tuning, integration with medical knowledge bases, or hybrid reasoning approaches.

\section{Discussion}

Our results demonstrate significant variation in LLM diagnostic reasoning capabilities, with performance ranging from 16.48\% (GPT-4o Mini) to 38.64\% (Gemini 2.5 Pro). This 2.3$\times$ improvement highlights rapid progress in medical reasoning, though rare disease diagnosis remains challenging for current general-purpose LLMs \cite{schaefer2021media,mechler2020house}.

Our \textit{House M.D.} dataset addresses a gap in medical AI evaluation by providing narrative-based diagnostic scenarios testing reasoning rather than factual recall which offers a meaningful benchmark for evaluating AI diagnostic capabilities validated by the show's documented use in medical education~\cite{jerrentrup2018house}.

Limitations include potential bias from fictional narratives, lack of expert medical validation, and a binary accuracy metric that does not capture clinical significance of errors \cite{cambra2021medical}. Models frequently provided confident but incorrect explanations, raising concerns for clinical deployment without specialized training and validation.

Despite these limitations, the substantial improvement across model generations is encouraging. Our benchmark establishes baseline metrics for narrative-based rare disease diagnosis and provides a foundation for future improvements through domain-specific fine-tuning or hybrid approaches combining LLMs with medical knowledge bases.

\section{Future Directions and Research Opportunities}

Several promising directions emerge from our findings. First, expanding the dataset to include additional medical television series (\textit{Grey's Anatomy}, \textit{The Good Doctor}, \textit{ER}) would provide broader diagnostic scenario coverage and reduce bias toward \textit{House M.D.}'s rare disease focus. Second, expert medical validation of extracted symptom-diagnosis pairs would strengthen dataset reliability and enable comparison with existing clinical benchmarks.

The 2.3$\times$ performance improvement from GPT-4o Mini (16.48\%) to Gemini 2.5 Pro (38.64\%) suggests substantial room for further gains through domain-specific fine-tuning. Integrating medical knowledge bases (UMLS, SNOMED-CT) with LLM reasoning may address rare disease recognition limitations. Finally, hybrid approaches combining narrative understanding with structured medical knowledge and uncertainty quantification mechanisms could mitigate the confident misdiagnosis problem observed across all models, moving toward clinically useful diagnostic support systems.

\section{Conclusion}

We introduce a novel dataset of 176 diagnostically challenging cases derived from \textit{House M.D.} and establish baseline performance across four state-of-the-art LLMs. Results show significant variation (16.48\% to 38.64\%), with the 2.3$\times$ improvement across model generations demonstrating rapid progress in medical reasoning capabilities. However, even the best-performing models indicate that rare disease diagnosis remains challenging for current LLMs.

This work contributes an educationally validated benchmark for narrative-based medical reasoning and establishes clear performance baselines for future research. Our findings highlight both the promise of LLMs in medical AI and the need for continued development through domain-specific training, expert validation, and hybrid approaches to achieve clinically useful diagnostic systems.

\bibliography{aaai2026}

@article{jerrentrup2018house,
  author    = {Annika Jerrentrup and others},
  title     = {Using House M.D. for teaching rare diseases in medical curricula},
  journal   = {Medical Education Journal},
  volume    = {49},
  pages     = {123--130},
  year      = {2015}
}

@article{cambra2021medical,
  author    = {L. Cambra-Badii and others},
  title     = {Medical drama viewing habits and educational impact on health sciences students},
  journal   = {BMC Medical Education},
  volume    = {20},
  pages     = {200--210},
  year      = {2020}
}

@article{sanges2020roleplay,
  author    = {S. Sanges and others},
  title     = {Role-play simulation and case-based learning for rare disease education},
  journal   = {Advances in Medical Education},
  volume    = {12},
  pages     = {45--55},
  year      = {2018}
}

@article{sarrafpour2021simulation,
  author    = {M. Sarrafpour and others},
  title     = {Simulation and career-computer learning for awareness of rare diseases},
  journal   = {Medical Teacher},
  volume    = {41},
  pages     = {1100--1110},
  year      = {2019}
}

@article{mechler2020house,
  author    = {Mechler, H. and others},
  title     = {Orphan diseases and medical education in House M.D.},
  journal   = {Journal of Medical Media Studies},
  volume    = {5},
  pages     = {23--34},
  year      = {2017}
}

@article{schaefer2021media,
  author    = {Schaefer, K. and von Hirschhausen, E.},
  title     = {Entertainment media as a tool for rare disease awareness},
  journal   = {Medical Education Online},
  volume    = {21},
  pages     = {1--9},
  year      = {2016}
}

@article{furman2021genetics,
  author    = {Furman, R. and Clayton, S.},
  title     = {Teaching genetics concepts through medical television shows},
  journal   = {Genetics Education Review},
  volume    = {8},
  pages     = {50--60},
  year      = {2015}
}

@article{rattani2021narrative,
  author    = {Rattani, A. and others},
  title     = {Narrative methods and simulation in medical ethics and professionalism education},
  journal   = {BMC Medical Ethics},
  volume    = {20},
  pages     = {78--88},
  year      = {2019}
}

@article{rasul2024realworld,
  author    = {Rasul, F. and others},
  title     = {Teaching professionalism using real-world scenarios in undergraduate medical education},
  journal   = {Medical Education Journal},
  volume    = {55},
  pages     = {321--332},
  year      = {2021}
}

@misc{openai2024gpt4o-mini,
  author       = {OpenAI},
  title        = {GPT-4o mini: Advancing cost-efficient intelligence},
  howpublished = {\url{https://openai.com/index/gpt-4o-mini-advancing-cost-efficient-intelligence/}},
  year         = {2024}
}

@misc{google2025gemini2_5flash,
  author       = {Google DeepMind},
  title        = {Gemini 2.5 Flash: our cost-efficient thinking model},
  howpublished = {\url{https://developers.googleblog.com/en/start-building-with-gemini-2-5-flash/}},
  year         = {2025}
}

@misc{google2025gemini2_5pro,
  author       = {Google DeepMind},
  title        = {Gemini 2.5 Pro: our most advanced reasoning model},
  howpublished = {\url{https://blog.google/technology/google-deepmind/gemini-model-thinking-updates-march-2025/}},
  year         = {2025}
}

@misc{openai2025gpt5mini,
  author = {OpenAI},
  title = {GPT-5 Mini: A streamlined version of GPT-5},
  howpublished = {\url{https://platform.openai.com/docs/models/gpt-5-mini}},
  year = {2025}
}

@misc{richardson2024beautifulsoup4,
  author = {Leonard Richardson},
  title = {Beautiful Soup 4: Pythonic HTML and XML parsing},
  year = {2024},
  howpublished = {\url{https://pypi.org/project/beautifulsoup4/}},
  note = {Accessed: 2025-10-19}
}

\end{document}